\documentclass[11pt]{article}

\usepackage[final]{acl}

\usepackage{times}
\usepackage{latexsym}

\usepackage[T1]{fontenc}

\usepackage[utf8]{inputenc}

\usepackage{microtype}

\usepackage{inconsolata}

\usepackage{graphicx}
\usepackage{amsmath}     
\usepackage{amssymb}     
\usepackage{amsfonts}    
\usepackage{booktabs}    
\usepackage{multirow}    
\usepackage{stfloats}
\usepackage{float}  
\title{Resonant Context Anchoring: Decoupling Attention Routing and Signal Gain at Inference Time}

\author{%
	\textbf{Mingkuan Zhao\textsuperscript{1}}\thanks{These authors contributed equally to this work.},
	\textbf{Yide Gao\textsuperscript{1}}\footnotemark[1],
	\textbf{Wentao Hu\textsuperscript{1}},
	\textbf{Suquan Chen\textsuperscript{1}},
	\textbf{Tianchen Huang\textsuperscript{2}}\thanks{Corresponding author.},
	\\
	\textbf{Zhenhua An\textsuperscript{1}},
	\textbf{Zetao Chang\textsuperscript{3}},
	\textbf{Xiayu Sun\textsuperscript{1}},
	\textbf{Yuheng Min\textsuperscript{4}}
	\\[0.8em]
	\parbox{\linewidth}{\small\centering
		\textsuperscript{1}Xi'an Jiaotong University,
		\textsuperscript{2}University of Science and Technology of China,\\
		\textsuperscript{3}Tongji University,
		\textsuperscript{4}Tsinghua University
	}
	\\[0.6em]
	\parbox{\linewidth}{\small\centering
		\texttt{\{mingkuanzhao, yidegao, wentao\_hu, suquanchen, zhenhuan, xiayu\_sun\}@stu.xjtu.edu.cn},\\
		\texttt{tchuang@mail.ustc.edu.cn},
		\texttt{2534034@tongji.edu.cn},
		\texttt{minyh24@mails.tsinghua.edu.cn}
	}
}

\begin{document}
\maketitle
\begin{abstract}
	Large Language Models (LLMs) frequently exhibit ``contextual disregard'' when faced with input evidence that conflicts with their internal parametric memory, leading to persistent factual hallucinations. Existing mitigation strategies primarily rely on suppressing specific neuron activations or employing computationally expensive contrastive decoding mechanisms, which often result in increased perplexity or significantly elevated inference latency. To address these limitations, we propose \textbf{Resonant Context Anchoring (RCA)}, a lightweight inference-time intervention method grounded in the perspective of residual stream signal dynamics. RCA aims to resolve the signal attenuation of external evidence during its propagation through deep networks. The core mechanism involves the orthogonal decoupling of routing logic and information magnitude within the self-attention module. By utilizing raw pre-softmax attention scores as an instantaneous metric of semantic alignment, we construct a dynamic gain field via non-linear rectification to selectively amplify the norms of value vectors corresponding to context tokens, without altering the attention probability distribution. This mechanism effectively elevates the signal-to-noise ratio (SNR) of input evidence within the residual stream mixture, thereby robustly anchoring the generation trajectory to the truthful context during inference. Extensive experiments on the Llama-3 model series demonstrate that RCA significantly improves contextual faithfulness across multiple factual consistency and strong knowledge-conflict tasks, effectively suppressing parametric hallucinations. Furthermore, results confirm that as a training-free and computationally negligible plug-and-play module, RCA achieves a Pareto improvement in faithfulness and fluency while maintaining the model's general language understanding capabilities. Our code is available at \url{https://github.com/yidGao/RCA-Implementation}.
\end{abstract}

\section{Introduction}

Despite the remarkable generalization capabilities demonstrated by Large Language Models (LLMs) 
\citep{NEURIPS2020_1457c0d6, openai2024gpt4technicalreport, he2024telechattechnicalreport, 
	li2024teleflmtechnicalreport} in natural language generation tasks, their reliance on parametric 
memory—internalized during pre-training—renders them susceptible to factual errors, particularly 
when processing input evidence that contradicts internal priors. This phenomenon, termed parametric hallucination \citep{10.1145/3571730}, reveals a fundamental tension between a model's internalized beliefs and the provided context. Recent scholarship indicates a persistent robustness failure: even when presented with accurate and relevant evidence, LLMs frequently exhibit ``contextual disregard,'' prioritizing intrinsic parametric priors over external information. This tendency to override input facts with internal biases significantly undermines the reliability of LLMs in knowledge-intensive and high-stakes reasoning scenarios.

Current approaches to mitigating these hallucinations predominantly fall into two categories: supervised fine-tuning and inference-time intervention. While Supervised Fine-Tuning (SFT) can improve instruction following through targeted data alignment, it incurs high computational costs and risks overfitting to specific conflict patterns, thereby limiting generalizability. Conversely, inference-time interventions attempt to modulate generation behavior without updating model weights. Contrastive Decoding (CD) strategies, for instance, rectify output distributions by contrasting logits generated with and without context. However, these methods typically necessitate multiple forward passes, doubling inference latency and rendering them impractical for real-time applications. Alternative methods based on Activation Engineering seek to identify and suppress specific attention heads or neurons responsible for retrieving parametric memory. Yet, such coarse-grained suppression often disrupts the underlying syntactic structure and semantic coherence, resulting in increased perplexity and degraded linguistic fluency.

\begin{figure}[H] 
	\centering
	\includegraphics[width=0.95\linewidth]{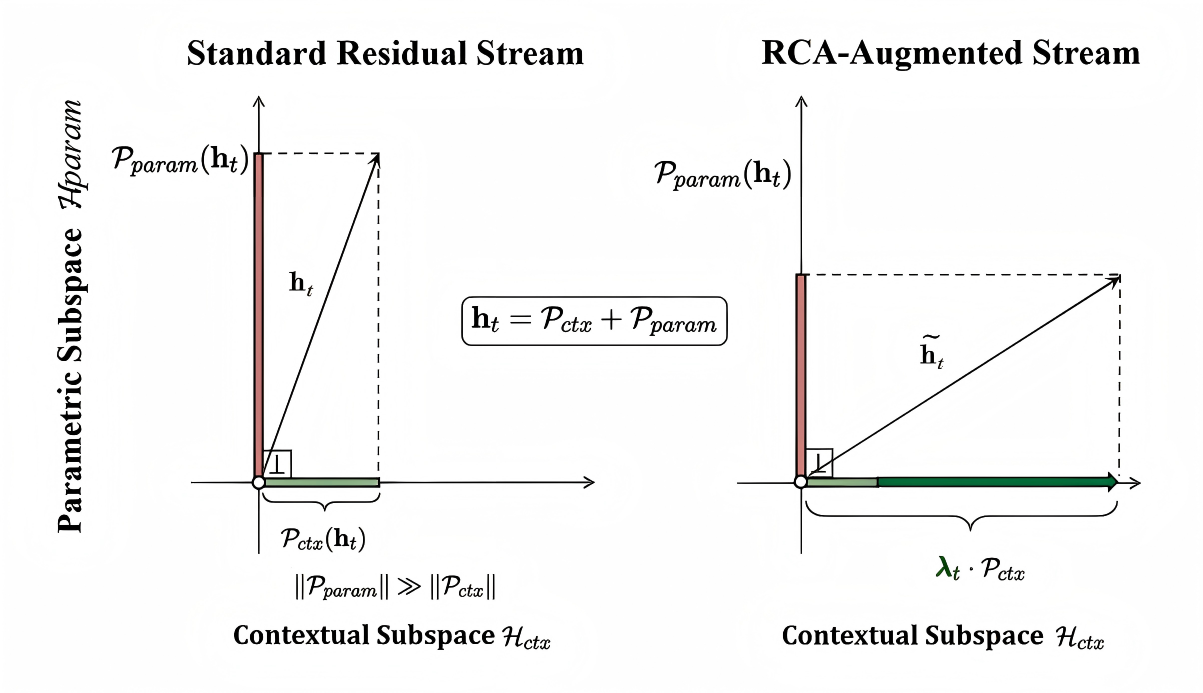}
	\caption{\textbf{Geometric interpretation of subspace projection.} 
		Left: In the standard residual stream, the parametric projection $\mathcal{P}_{param}$ dominates the energy, causing the state $\mathbf{h}_t$ to align with internal priors. 
		Right: RCA applies a gain $\lambda_t$ to the contextual projection, effectively rotating the resultant vector $\tilde{\mathbf{h}}_t$ towards the contextual manifold without altering the underlying basis.}
	\label{fig:projection_dynamics}
\end{figure}

In this work, we revisit the phenomenon of contextual disregard through the lens of signal dynamics within the residual stream. We hypothesize that the model's failure in conflict scenarios stems not from an inability to identify relevant context (a routing error), but from the insufficient energy of the contextual signal during propagation through deep networks (a gain deficit). Within the Transformer architecture \citep{NIPS2017_3f5ee243}, the residual stream can be modeled as a linear superposition of parametric and contextual subspaces. Even if the query vector correctly attends to context tokens, if the norms of the corresponding Value vectors are significantly smaller than the parametric noise vectors injected by the Multi-Layer Perceptrons (MLPs)\citep{geva2021transformerfeedforwardlayerskeyvalue}, the effective information becomes submerged, leading to a critical drop in the Signal-to-Noise Ratio (SNR). Consequently, the solution to factual anchoring lies not in the destructive suppression of internal memory, but in the selective amplification of the contextual signal magnitude. This conceptual shift in geometric projection is illustrated in Figure \ref{fig:projection_dynamics}.

To this end, we propose \textbf{Resonant Context Anchoring (RCA)}, a lightweight inference-time intervention method. The core contribution of RCA is the orthogonal decoupling of ``routing decisions'' and ``information gain'' within the self-attention mechanism. Specifically, we utilize raw pre-softmax attention scores as an instantaneous metric of semantic alignment to drive a non-linear rectification mechanism. During inference, when a high semantic correlation (resonance) between the query and context tokens is detected, RCA dynamically amplifies the norms of the corresponding Value vectors. This process significantly enhances the weight of input evidence within the residual stream without altering the attention probability distribution, thereby preserving both syntactic correctness and factual faithfulness.

We conduct extensive evaluations on the Llama-3 model series. Empirical results demonstrate that RCA significantly improves contextual faithfulness in summarization and strong knowledge-conflict benchmarks (e.g., NQ-Swap), effectively mitigating hallucinations driven by prior bias. Crucially, as a training-free plug-and-play module introducing only negligible element-wise operations, RCA achieves a Pareto improvement in faithfulness and fluency without incurring the latency penalties associated with iterative decoding methods.

\section{Related Work}

Contextual grounding has emerged as a fundamental framework for mitigating factual hallucinations in Large Language Models (LLMs) by anchoring outputs in external evidence \cite{lewis2021retrievalaugmentedgenerationknowledgeintensivenlp}. However, a persistent challenge arises when input evidence contradicts the model's internalized parametric memory, leading to ``contextual disregard'' where the model prioritizes intrinsic priors over provided facts \cite{longpre-etal-2021-entity}. Previous scholarship has highlighted the limitations of LLMs in handling complex conflicting evidence and long-tail knowledge \cite{chen2022richknowledgesourcesbring, mallen-etal-2023-trust}, necessitating adaptive augmentation strategies that dynamically calibrate the reliance on external context based on generation confidence \cite{wang-etal-2025-adaptive}. These findings underscore the need for decoding mechanisms that robustly anchor generation to the input context during knowledge conflicts.

The foundation of the hallucination problem lies in the pre-training paradigm of modern LLMs. 
Models such as the TeleChat series \citep{wang-etal-2024-telechat, he2024telechattechnicalreport, 
	wang2025technicalreporttelechat2telechat25, liu2025trainingreporttelechat3moe} and the 
Tele-FLM family \citep{li2024teleflmtechnicalreport, li202452b1tlessonslearned} demonstrate 
that bilingual and large-scale pre-training over trillions of tokens endows models with strong 
generalization, yet simultaneously entrenches parametric priors that may conflict with externally 
provided evidence at inference time. These capable base models have been deployed across a broad 
spectrum of knowledge-intensive downstream tasks: TableReasoner 
\citep{xiong2025tablereasoneradvancingtablereasoning} advances structured table reasoning by 
leveraging LLMs' semantic understanding, MR-UIE \citep{li2025mruiemultiperspectivereasoningreinforcement} 
applies multi-perspective reinforcement learning to universal information extraction, and 
\citet{zhao2025enhancing} enhance mathematical reasoning through computation logic graphs. 
The LLMSR pipeline \citep{xing-etal-2025-llmsr} further addresses the challenge of structured 
reasoning data construction, combining gradient-reward policy optimization with iterative 
self-refinement. Collectively, these works underscore that while LLMs can reliably internalize 
and apply broad world knowledge, their tendency to override contextual evidence with parametric 
memory represents a critical barrier to deployment in high-reliability, knowledge-conflict scenarios.

Complementary to faithfulness research, recent work has revisited the internal routing logic of 
the attention mechanism from both efficiency and structural perspectives. 
\citet{zhao2025makingheadcountsparse} propose a sparse attention strategy that achieves 
competitive performance without sacrificing inference speed, demonstrating that not all attention 
heads contribute equally to the information routing process. In parallel, Mosaic Pruning 
\citep{hu2025mosaicpruninghierarchicalframework} introduces a hierarchical pruning framework for 
Mixture-of-Experts (MoE) models that generalizes across model families, revealing that structured 
sparsity in attention and feed-forward blocks can be exploited for efficient deployment without 
significant performance loss. These findings are closely aligned with the signal-level 
perspective adopted in our work: if attention routing can be decoupled from signal magnitude 
for the purpose of efficiency, the same decoupling principle can be leveraged to selectively 
amplify contextual evidence signals during inference—the central insight behind RCA.

Inference-time intervention strategies represent a primary category of techniques for enhancing model faithfulness. Contrastive Decoding (CD) optimizes reasoning quality by calibrating output distributions across different model configurations \cite{obrien2023contrastivedecodingimprovesreasoning}. Notably, Context-aware Decoding (CAD) enforces context adherence by amplifying the logit difference between conditional and unconditional generations \cite{shi-etal-2024-trusting}. Other methods, such as DoLa, contrast mature and premature layer representations to improve factuality \cite{chuang2024doladecodingcontrastinglayers}, while Inference-Time Intervention (ITI) injects linear directions associated with truthfulness into specific attention heads \cite{li2024inferencetimeinterventionelicitingtruthful}. Although effective, these approaches often introduce significant computational overhead or require the training of auxiliary probes.

Advances in activation engineering and mechanistic interpretability offer a physical basis for precisely steering model behavior \cite{turner2024steeringlanguagemodelsactivation}. Research has demonstrated that latent steering vectors can be extracted to manipulate the semantic space of pretrained decoders without fine-tuning \cite{subramani-etal-2022-extracting}. Techniques like Contrastive Activation Addition (CAA) apply displacement vectors in the residual stream to guide specific behaviors such as factuality \cite{rimsky-etal-2024-steering}. Structurally, feed-forward layers are identified as key-value memories for parametric facts \cite{geva2021transformerfeedforwardlayerskeyvalue}, while induction heads facilitate the replication of contextual information across layers \cite{olsson2022incontextlearninginductionheads}. Recent mathematical perspectives on Transformer dynamics \cite{geshkovski2025mathematicalperspectivetransformers} and methods for editing implicit assumptions \cite{orgad2023editingimplicitassumptionstexttoimage} further support the feasibility of fine-grained signal modulation within the residual stream.

The maturation of factuality evaluation frameworks provides quantitative metrics for assessing intervention efficacy. FACTScore enables fine-grained evaluation through atomic fact decomposition \cite{min-etal-2023-factscore}, while AlignScore utilizes a unified alignment function to measure factual consistency across diverse tasks \cite{zha-etal-2023-alignscore}. These metrics, combined with systematic analyses of LLMs as factual reasoners \cite{laban2023llmsfactualreasonersinsights}, form the experimental foundation for modern research. Building upon these foundations, our proposed RCA method introduces a lightweight intervention that recalibrates the signal-to-noise ratio in the residual stream, offering an efficient alternative to contrastive decoding while preserving architectural integrity better than coarse-grained activation suppression.

\section{Methodology}
\label{sec:method}

In this section, we formally define the algorithmic framework of \textbf{Resonant Context Anchoring (RCA)}. We reconstruct the self-attention mechanism of the Transformer architecture as a signal transmission system equipped with adaptive spectral gain control. Based on the theory of vector space decomposition within the residual stream, we derive a non-linear rectification mechanism driven by semantic correlation. This approach aims to dynamically calibrate the energy distribution of the contextual subspace by orthogonally decoupling the routing decision logic from the signal intensity.

\subsection{Subspace Decomposition and SNR Dynamics in Residual Streams}
\begin{figure}[H]
	\centering
	\includegraphics[width=\linewidth]{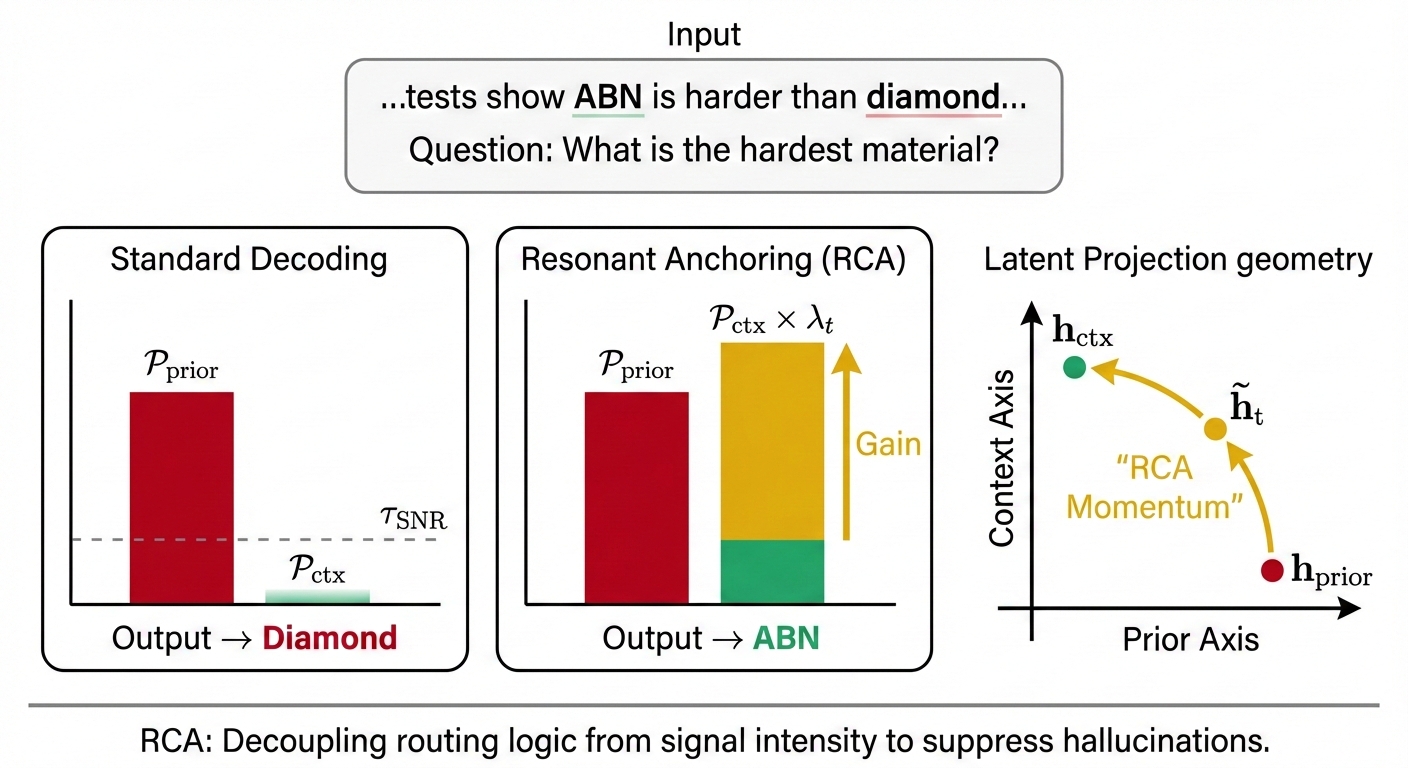}
	\caption{\textbf{Mechanistic illustration of signal dynamics.} 
		Left: Standard decoding fails when contextual evidence is submerged by priors. 
		Middle: RCA restores the signal-to-noise ratio. 
		Right: The resulting "RCA Momentum" in latent space.}
	\label{fig:toy_example}
\end{figure}

\begin{figure*}[h]  
	\centering
	\includegraphics[width=0.95\textwidth]{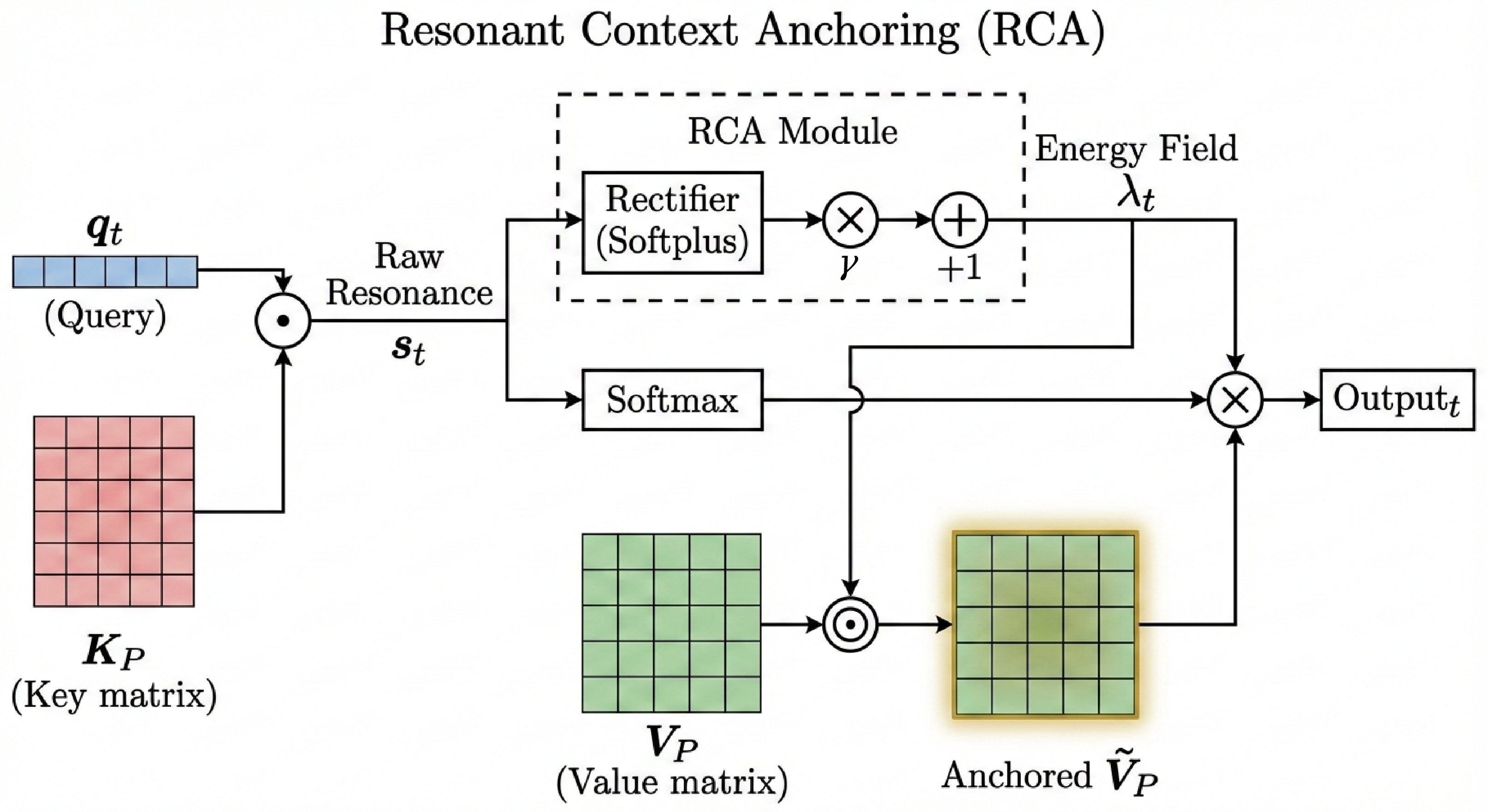} 
	\caption{\textbf{The architecture of Resonant Context Anchoring (RCA).} 
		The module decouples routing logic (Softmax) and information magnitude. 
		It computes an energy field $\lambda_t$ via a non-linear rectifier (Softplus) applied to the raw resonance $s_t$, 
		which is then used to modulate the value matrix $V_P$ to produce the anchored representation $\tilde{V}_P$.}
	\label{fig:rca_architecture}
\end{figure*}

At the $l$-th layer of the Transformer architecture, the hidden state vector $\mathbf{h}_t^{(l)} \in \mathbb{R}^d$ can be formalized as a linear superposition of distinct information sources \citep{elhage2021mathematical} within a high-dimensional latent space. For the $t$-th time step in an autoregressive generation process, the state update follows the residual connection rule, driven jointly by the Multi-Head Attention (MHA) and Feed-Forward Network (FFN) sub-layers. We decompose this state space $\mathbb{R}^d$ into two competing subspaces: the Contextual Subspace $\mathcal{H}_{ctx} \subset \mathbb{R}^d$, spanned by external evidence within the current context window, and the Parametric Subspace $\mathcal{H}_{param} \subset \mathbb{R}^d$, implicitly encoded by the model's pre-trained weight matrices. Consequently, the residual vector $\mathbf{h}_t$ can be expressed as the superposition of two orthogonal projection components:
\begin{equation}
	\mathbf{h}_t \approx \mathcal{P}_{ctx}(\mathbf{h}_t) + \mathcal{P}_{param}(\mathbf{h}_t)
\end{equation}
where $\mathcal{P}_{ctx}$ and $\mathcal{P}_{param}$ denote the projection operators onto the contextual and parametric subspaces, respectively. The mechanism underlying hallucinations or knowledge conflicts can be interpreted from a signal processing perspective as a critical imbalance in the Signal-to-Noise Ratio (SNR) within the residual stream, leading to what we term ``signal submergence'' \textbf{(Figure \ref{fig:toy_example})}. Specifically, when processing a specific query $\mathbf{q}_t$, if the $L_2$ norm of the parametric component significantly dominates the residual stream, satisfying the inequality condition:
\begin{equation}
	\|\mathcal{P}_{param}(\mathbf{h}_t)\|_2 \gg \|\mathcal{P}_{ctx}(\mathbf{h}_t)\|_2
\end{equation}

the model's generation trajectory will collapse onto the internal prior manifold, leading to the disregard of external evidence. The standard attention mechanism employs the Softmax function to achieve probabilistic normalization; however, this operation only constraints the routing distribution of information aggregation without explicitly controlling the scalar magnitude of the aggregated information. Even if the attention weights correctly point to context tokens, if the raw magnitude of the value vector $\mathbf{v}_i$ is insufficient to counteract the high-norm parametric noise injected by the FFN layers in deep networks, the effective signal will attenuate during hierarchical propagation. Therefore, our objective is to construct a gain operator $\mathcal{G}(\cdot)$ that selectively enhances the modified contextual projection norm $\|\mathcal{P}_{ctx}(\mathcal{G}(\mathbf{h}_t))\|$ while maintaining the invariant routing distribution $\alpha_{t} \in \Delta^{L-1}$.

\subsection{Semantic Spectrum Analysis and In-Situ Value Modulation}

The detailed internal architecture of the RCA module, which implements the decoupling of routing logic and signal gain, is depicted in \textbf{Figure \ref{fig:rca_architecture}}. 

To achieve the aforementioned gain objective, RCA introduces a non-linear modulation mechanism based on the semantic alignment spectrum. This mechanism operates directly on the value projection stage of the attention layer, realizing the orthogonal decoupling of routing logic and signal gain. First, for the $h$-th attention head, we compute the raw correlation score between the current query vector $\mathbf{q}_t \in \mathbb{R}^{d_k}$ and the context key vector matrix $\mathbf{K} \in \mathbb{R}^{L \times d_k}$. This unnormalized dot product constitutes the raw spectrum distribution of semantic alignment $\mathbf{s}_t \in \mathbb{R}^L$, where the $i$-th component is defined as:
\begin{equation}
	s_{t,i} = \frac{\mathbf{q}_t^\top \mathbf{k}_i}{\sqrt{d_k}}
\end{equation}
This scalar precisely quantifies the degree of collinearity between the query intent and the $i$-th context token in the semantic space. Unlike the standard attention mechanism which utilizes $\mathbf{s}_t$ solely for computing Softmax probabilities, RCA introduces a parallel rectification path to construct a dynamic gain field. We define a non-linear rectification function $\psi: \mathbb{R} \to \mathbb{R}^+$ to extract the "resonance intensity." To ensure the smoothness of the manifold and avoid gradient truncation, we employ the Softplus operator with a sensitivity coefficient $\gamma$, constructing a scalar gain $\lambda_{t,i}$ for each token:
\begin{equation}
	\lambda_{t,i} = 1 + \gamma \cdot \psi(s_{t,i}) = 1 + \gamma \cdot \ln(1 + \exp(s_{t,i}))
\end{equation}
The gain function $\lambda_{t,i}$ possesses rigorous mathematical properties: as $s_{t,i} \to -\infty$ (indicating semantic orthogonality or negative correlation), $\lambda_{t,i} \to 1$, causing the system to degenerate into an identity mapping and ensuring zero interference with noise; when $s_{t,i} > 0$ (indicating semantic resonance), $\lambda_{t,i}$ exhibits quasi-linear monotonic growth with respect to correlation. Subsequently, we apply this gain field to the value vector matrix $\mathbf{V} \in \mathbb{R}^{L \times d_v}$ via an in-situ Hadamard product modulation. The modified value vector $\tilde{\mathbf{v}}_i$ is defined as:
\begin{equation}
	\tilde{\mathbf{v}}_i = \lambda_{t,i} \cdot \mathbf{v}_i
\end{equation}
Finally, the output of the attention head $\mathbf{o}_t$ is computed via the standard probability-weighted summation, but taking the energy-enhanced value vectors as input. The complete computational form can be expanded as:
\begin{equation}
	\begin{split}
		\mathbf{o}_t = \sum_{i=1}^{L} \text{Softmax}(\mathbf{s}_t)_i \cdot \\
		\left[ (1 + \gamma \cdot \text{Softplus}(s_{t,i})) \cdot \mathbf{v}_i \right]
	\end{split}
\end{equation}
From the perspective of vector synthesis, this operation is equivalent to applying a momentum term controlled jointly by $\gamma$ and $s_{t,i}$ in the direction of the context-relevant gradient, forcing the residual stream to converge towards the ground-truth evidence manifold.

\subsection{Asymptotic Analysis of Local SNR Gain}

To theoretically validate the effectiveness of RCA, we perform an asymptotic analysis on the gain boundary of the local Signal-to-Noise Ratio (SNR). Let the effective signal energy $E_{sig}$ at step $t$ be the weighted norm of value vectors within the relevant context set $\mathcal{I}_{rel}$, and the noise energy $E_{noise}$ be the norm of irrelevant context and parametric bias. Under the standard attention mechanism, the base SNR is given by $\text{SNR}_{base} = E_{sig} / E_{noise}$. Upon introducing RCA, due to the non-negativity and monotonicity of the rectification function $\psi$, for any $i \in \mathcal{I}_{rel}$, the high alignment with the query vector implies $s_{t,i} > 0$, thereby ensuring $\lambda_{t,i} > 1$. The modified signal energy $\tilde{E}_{sig}$ can be expressed as:
\begin{equation}
	\tilde{E}_{sig} = \left\| \sum_{i \in \mathcal{I}_{rel}} \alpha_{t,i} (1 + \gamma \psi(s_{t,i})) \mathbf{v}_i \right\|
\end{equation}
Utilizing the triangle inequality and its reverse property, and assuming the resonance intensity distribution within the relevant context is concentrated, we have $\tilde{E}_{sig} \approx (1 + \gamma \bar{\psi}) E_{sig}$, where $\bar{\psi}$ denotes the average resonance intensity of relevant tokens. Meanwhile, since RCA does not intervene in the FFN layer outputs, and for irrelevant tokens $j \notin \mathcal{I}_{rel}$, $s_{t,j}$ is typically negative leading to $\lambda_{t,j} \approx 1$, the noise energy term remains quasi-constant, i.e., $\tilde{E}_{noise} \approx E_{noise}$. Consequently, the post-intervention SNR satisfies the following inequality:
\begin{equation}
	\begin{split}
		\text{SNR}_{RCA} &\approx \frac{(1 + \gamma \bar{\psi}) E_{sig}}{E_{noise}} \\
		&= (1 + \gamma \bar{\psi}) \text{SNR}_{base} > \text{SNR}_{base}
	\end{split}
\end{equation}
This inequality mathematically guarantees that RCA monotonically enhances the model's sensitivity to external evidence. By adjusting the hyperparameter $\gamma$, we can explicitly control the gain magnitude of the SNR, thereby effectively escaping the hallucination attractor without disrupting the model's language modeling distribution.

\section{Experiments}
\label{sec:experiments}

In this section, we present a comprehensive empirical evaluation to assess the effectiveness, robustness, and parametric sensitivity of Resonant Context Anchoring (RCA). Our experimental design aims to systematically address three core research questions: First, can the proposed method significantly enhance contextual faithfulness in summarization tasks without compromising generation quality? Second, in adversarial scenarios characterized by strong conflicts between parametric memory and external evidence, does the method effectively suppress hallucinations and enforce context adherence? Third, does the method exhibit general safety, maintaining the model's fundamental capabilities in tasks lacking relevant context? All experiments are conducted on the Llama-3-8B-Instruct and Llama-3-70B-Instruct models \citep{grattafiori2024llama3herdmodels}, benchmarking RCA against standard greedy decoding strategies.

\subsection{Experimental Setup and Benchmarks}

To rigorously evaluate model performance across varying cognitive loads, we constructed a diverse suite of benchmarks encompassing single-document summarization, knowledge-conflict QA, and general commonsense QA. For the assessment of contextual faithfulness, we utilized the XSum dataset \citep{narayan-etal-2018-dont}, employing FactKB \citep{feng-etal-2023-factkb} and AlignScore \citep{zha-etal-2023-alignscore} as primary metrics for factual consistency between the generated text and source documents, complemented by ROUGE-L \citep{lin-2004-rouge} to evaluate content coverage. In the more challenging domain of knowledge conflict, we selected the NQ-Swap\citep{longpre-etal-2021-entity} and MemoTrap\citep{mckenzie2024inversescalingbiggerisnt} datasets. NQ-Swap constructs counter-factual contexts by substituting key entities (e.g., swapping countries or capitals) within Wikipedia passages, forcing the model to choose between internal correct knowledge and external altered context; we report Exact Match (EM) as the core metric for context adherence. MemoTrap tests the model's ability to resist rote memorization when completing counter-intuitive proverbs. For general capability evaluation, we employed the closed-book settings of TriviaQA \citep{joshi-etal-2017-triviaqa} and TruthfulQA \citep{lin-etal-2022-truthfulqa} to verify that the inference-time intervention does not induce catastrophic forgetting of general world knowledge. Implementation-wise, RCA is integrated as a parameter-free module within the Hugging Face Transformers library\citep{wolf-etal-2020-transformers}, with inference performed on NVIDIA A100 GPU clusters.

\subsection{Contextual Faithfulness and Conflict Resolution}

\begin{table*}[t]
	\centering
	\caption{\textbf{Main Results on Faithfulness and Knowledge Conflict.} The table reports the performance of standard greedy decoding (Baseline) versus RCA across two model scales. RCA demonstrates consistent improvements across all metrics. Notably, it achieves a Pareto improvement by enhancing faithfulness (FactKB, AlignScore) without degrading content coverage (ROUGE-L). ($\uparrow$ indicates higher is better).}
	\label{tab:main_results}
	\setlength{\tabcolsep}{9pt}
	\begin{tabular}{llcccc}
		\toprule
		\multirow{2}{*}{\textbf{Task}} & \multirow{2}{*}{\textbf{Metric}} & \multicolumn{2}{c}{\textbf{Llama-3-8B}} & \multicolumn{2}{c}{\textbf{Llama-3-70B}} \\
		\cmidrule(lr){3-4} \cmidrule(lr){5-6}
		& & Baseline & \textbf{RCA} & Baseline & \textbf{RCA} \\
		\midrule
		\multicolumn{6}{l}{\textit{\textbf{Contextual Faithfulness (XSum)}}} \\
		\multirow{3}{*}{Summarization} & FactKB ($\uparrow$) & 47.61 & \textbf{50.14} & 61.32 & \textbf{61.72} \\
		& AlignScore ($\uparrow$) & 58.20 & \textbf{59.45} & 65.10 & \textbf{65.88} \\
		& ROUGE-L ($\uparrow$) & 19.90 & \textbf{20.03} & 22.41 & 22.21 \\
		\midrule
		\multicolumn{6}{l}{\textit{\textbf{Knowledge Conflict Resolution}}} \\
		NQ-Swap & Exact Match ($\uparrow$) & 60.62 & \textbf{64.54} & 76.11 & \textbf{77.46} \\
		\multirow{2}{*}{MemoTrap} & Micro Acc ($\uparrow$) & 64.40 & \textbf{65.77} & 66.52 & \textbf{77.35} \\
		& Macro Acc ($\uparrow$) & 65.86 & \textbf{66.69} & 68.47 & \textbf{75.58} \\
		\bottomrule
	\end{tabular}
\end{table*}

The main experimental results, presented in Table \ref{tab:main_results}, indicate that RCA achieves performance significantly superior to the baseline in enhancing contextual faithfulness. In the XSum summarization task, the Llama-3-8B model with RCA shows an increase in the FactKB score from a baseline of 47.61 to 50.14, alongside a concurrent rise in AlignScore from 58.20 to 59.45. Crucially, this improvement in factual consistency is not attained at the expense of content richness; the ROUGE-L score remains stable or improves slightly (increasing from 19.90 to 20.03). Similar trends are observed in the larger Llama-3-70B model, where FactKB improves from 61.32 to 61.72 and AlignScore from 65.10 to 65.88. These findings corroborate that RCA does not merely circumvent hallucinations by truncating outputs, but rather successfully steers the model to attend more intensively to key information fragments within the source document, thereby realizing a Pareto improvement between faithfulness and informativeness.

In tasks characterized by severe conflicts between parametric memory and external evidence, RCA exhibits a robust capability to suppress internal priors. On the NQ-Swap dataset, the Llama-3-8B model under standard decoding achieves an Exact Match rate of only 60.62\%, implying that in approximately 40\% of cases, the model prioritizes internal correct facts over the specified context. The introduction of RCA boosts the Exact Match rate to 64.54\%, an absolute improvement of 3.92\%. This trajectory is maintained in the Llama-3-70B model, where the Exact Match rate rises from 76.11\% to 77.46\%. Results from the MemoTrap dataset further substantiate this conclusion: for Llama-3-70B, the Micro Accuracy surges from 66.52\% to 77.35\%, and Macro Accuracy improves from 68.47\% to 75.58\%. These data provide compelling evidence that by dynamically amplifying contextual signals resonating with the query in the residual stream, RCA effectively breaks the hold of parametric attractors, enabling the model to anchor robustly to external evidence even amidst intense conflict.

\subsection{Parameter Sensitivity and Non-linear Dynamics}

\begin{table}[t]
	\centering
	
	\caption{\textbf{Sensitivity Analysis of $\gamma$ on NQ-Swap.} The Exact Match (EM) scores illustrate the non-linear dynamics. Performance peaks in the $[0.04, 0.05]$ interval ("sweet spot") and degrades significantly beyond $0.08$.}
	\label{tab:ablation_gamma}
	\setlength{\tabcolsep}{4pt} 
	\begin{tabular}{lcc}
		\toprule
		\textbf{$\gamma$ Setting} & \textbf{Llama-3-8B} & \textbf{Llama-3-70B} \\
		\midrule
		Baseline ($\gamma=0$) & 60.62 & 76.11 \\
		$\gamma=0.02$ & 62.15 & 76.85 \\
		$\gamma=0.04$ & \textbf{64.54} & 77.20 \\
		$\gamma=0.05$ & 64.10 & \textbf{77.46} \\
		$\gamma=0.08$ & 62.80 & 76.90 \\
		$\gamma=0.10$ & 58.40 & 73.50 \\
		$\gamma=0.12$ & 52.10 & 68.20 \\
		\bottomrule
	\end{tabular}
\end{table}

The resonance sensitivity coefficient $\gamma$ serves as the core hyperparameter controlling the intensity of the RCA intervention. We conducted a granular grid search and ablation analysis (detailed in Table \ref{tab:ablation_gamma}) to investigate its non-linear impact on model behavior. The experiments reveal a clear and consistent performance evolution trajectory. In the interval where $\gamma$ ranges from 0 to 0.04, all key metrics exhibit a monotonic upward trend as the coefficient increases. For instance, with Llama-3-8B on NQ-Swap, increasing $\gamma$ from 0.02 to 0.04 raises the Exact Match rate from 62.15\% to a peak of 64.54\%. The model performance reaches a global optimum plateau when $\gamma$ lies between 0.04 and 0.05. Specifically, Llama-3-8B performs best at $\gamma=0.04$, while Llama-3-70B peaks at $\gamma=0.05$ (77.46\%), a difference likely attributable to the larger model's greater embedding space norm, which necessitates slightly stronger external excitation to overcome internal priors.

However, precise physical boundaries exist for parameter settings. When $\gamma$ exceeds 0.05 and approaches 0.08, performance on select datasets begins to decline; for example, the Exact Match rate for Llama-3-8B regresses to 62.80\% at $\gamma=0.08$. As $\gamma$ further increases beyond 0.1, the perplexity of the generated text rises sharply, leading to severe degradation in output quality. In the extreme setting of $\gamma=0.12$, performance drops to 52.10\% for Llama-3-8B and 68.20\% for Llama-3-70B. Based on this empirical analysis, we fix the optimal configurations for Llama-3-8B and Llama-3-70B at 0.04 and 0.05, respectively, and establish 0.08 as the hard upper threshold for the safe application of this method. This result unveils a distinct "stable region" for the resonance mechanism, within which contextual control can be maximized without introducing adverse side effects.

\subsection{Safety and Harmlessness Evaluation}

\begin{table}[h]
	\centering
	\caption{\textbf{Safety Evaluation on Closed-Book Tasks.} We report performance on tasks with irrelevant or no context provided. The negligible deviation between Baseline and RCA confirms the method's harmlessness and robust general capabilities.}
	\label{tab:safety}
	\setlength{\tabcolsep}{8pt}
	\begin{tabular}{llcc}
		\toprule
		\textbf{Task} & \textbf{Metric} & \textbf{Baseline} & \textbf{RCA} \\
		\midrule
		\multirow{2}{*}{TruthfulQA} & MC1 & 38.92 & 38.92 \\
		& MC2 & 55.64 & 55.61 \\
		TriviaQA & EM & 56.58 & 56.52 \\
		PopQA & Acc & 26.64 & 26.59 \\
		\bottomrule
	\end{tabular}
\end{table}

To verify the harmlessness of RCA, we conducted general QA testing in settings where either no relevant context or irrelevant context was provided (results in Table \ref{tab:safety}). On the TruthfulQA MC1 (Multiple Choice-Single) metric, the model with RCA scores identically to the baseline (38.92\%), and on the MC2 (Multiple Choice-Multiple) metric, the difference is a negligible 0.03\% (55.64\% vs. 55.61\%). Similarly, on the TriviaQA dataset, the Exact Match rate fluctuates only marginally from 56.58\% to 56.52\%, well within the margin of statistical error. PopQA results exhibit the same trend, with accuracy shifting nominally from 26.64\% to 26.59\%. The theoretical basis for this result lies in RCA's resonance detection mechanism: in general QA scenarios, the dot product scores between the query vector and irrelevant or empty contexts naturally fall into low or negative ranges. After processing through the Softplus rectification, the amplification coefficient approaches 1, effectively rendering the intervention mechanism "dormant." This adaptive activation characteristic ensures that RCA can function as an always-on decoding strategy, enhancing performance on specific context-intensive tasks without necessitating concerns regarding negative impacts on the model's general linguistic capabilities.

\section{Limitations}

While Resonant Context Anchoring (RCA) demonstrates significant efficacy in suppressing parametric hallucinations and reinforcing adherence to input evidence, the introduction of the resonance sensitivity coefficient $\gamma$ necessitates calibration across different model scales or task environments. \textit{However}, empirical evidence indicates that this parameter exhibits cross-model stability, with optimal values remaining largely consistent between Llama-3-8B and 70B. This implies that parameter validation requires only a minimal development set, ensuring RCA imposes no significant tuning burden without compromising generalizability.

Regarding the theoretical framework, this study adopts a linear approximation perspective of subspace projections within the residual stream, which may simplify the highly non-linear dynamics inherent in Transformer architectures. \textit{Nevertheless}, this simplification represents an optimal trade-off between inference efficiency and intervention precision. The empirical success across multiple strong knowledge-conflict and factual consistency benchmarks provides compelling evidence for the validity of this theoretical assumption, demonstrating that the orthogonal decoupling logic is sufficient to capture critical signal gains and mitigate hallucinations. Furthermore, although the current evaluation focuses on faithfulness to provided textual evidence, the underlying logic of RCA operates on the universal self-attention mechanism. Consequently, its application is not restricted to specific task types, and preliminary safety evaluations confirm that the method induces no negative interference with the model's general language modeling capabilities.

Finally, while the primary experiments were conducted using the Llama-3 series, this does not imply an architectural dependence. Because RCA targets the core Query-Key-Value interaction logic and the energy distribution of the residual stream---rather than specific weight distributions or layer configurations---the method is theoretically architecture-agnostic and directly transferable to other autoregressive Transformer-based models. Moreover, as a training-free and computationally negligible plug-and-play intervention, RCA offers superior deployment advantages over traditional fine-tuning or high-latency contrastive decoding strategies, with its performance benefits far outweighing the minimal adaptation costs in specialized edge-case scenarios.

\section{Conclusion}
\label{sec:conclusion}
In this paper, we address the pervasive issue of parametric memory interference and factual hallucinations in context-grounded generation by proposing \textbf{Resonant Context Anchoring (RCA)}, an inference-time intervention framework grounded in the perspective of residual stream signal dynamics. Theoretically, RCA orthogonally decouples the routing logic from the signal intensity within the attention mechanism. Practically, it implements adaptive energy injection for contextual evidence via a non-linear rectification mechanism. Distinct from traditional suppressive interventions, RCA operates on the hypothesis of ``signal attenuation,'' advocating for the enhancement of the Signal-to-Noise Ratio (SNR) of effective information to counteract the parametric noise inherent in deep neural networks.

Through asymptotic analysis of residual stream dynamics and extensive empirical validation, we confirm that hallucinations frequently stem from the insufficient relative energy of correct contextual signals during propagation, rather than the model's inability to identify relevance. By dynamically amplifying the norms of Value vectors that align with query semantics during inference, RCA successfully reverses this signal imbalance, enabling input evidence to override internal prior biases and establish dominance. Experimental results on the Llama-3 model series demonstrate that RCA achieves consistent improvements significantly superior to baselines across multiple summarization and counterfactual conflict tasks, realizing a Pareto improvement in faithfulness and fluency. Furthermore, its zero-degradation performance on general closed-book QA tasks validates the method's high safety and robustness, qualifying it as a viable always-on decoding strategy.

Compared to computationally expensive iterative contrastive decoding or data-intensive supervised fine-tuning, RCA establishes a novel paradigm that is efficient, training-free, and fully interpretable. It demonstrates that precise physical modulation of internal activation states can correct cognitive biases in large language models without disrupting their pre-trained linguistic manifolds. This achievement not only provides an industrially viable solution for high-reliability generation systems but also opens new theoretical pathways for understanding the internal dynamics of Transformers from a signal processing perspective.

\bibliography{custom}

\appendix

\end{document}